%% file: lrec-coling2024-example.tex
\documentclass[10pt, a4paper]{article}
\usepackage{times}
\usepackage{latexsym}
\usepackage{graphicx}
\usepackage{xcolor}
\usepackage{array}
\usepackage{caption}
\usepackage{subcaption}
\usepackage{amsmath}
\usepackage{amsfonts}
\usepackage{amssymb}
\usepackage{bbm}
\usepackage{multirow}
\usepackage{float}
\usepackage{amsmath,amsfonts,amssymb}
\usepackage[]{lrec-coling2024} 
\newcommand\scalemath[2]{\scalebox{#1}{\mbox{\ensuremath{\displaystyle #2}}}}

\title{Query-driven Relevant Paragraph Extraction from Legal Judgments}

\name{Santosh T.Y.S.S, Elvin Quero Hernandez, Matthias Grabmair} 

\address{School of Computation, Information, and Technology; \\ Technical University of Munich, Germany
\\
\{santosh.tokala, elvin.quero, matthias.grabmair\}@tum.de\\}

\abstract{
Legal professionals often grapple with navigating lengthy legal judgements to pinpoint information that directly address their queries. This paper focus on this task of extracting relevant paragraphs from legal judgements based on the query. We construct a specialized dataset for this task from the European Court of Human Rights (ECtHR) using the case law guides. We assess the performance of current retrieval models in a zero-shot way and also establish fine-tuning benchmarks using various models. The results highlight the significant gap between fine-tuned and zero-shot performance, emphasizing the challenge of handling distribution shift in the legal domain. We notice that the legal pre-training handles distribution shift on the corpus side but still struggles on query side distribution shift, with unseen legal queries.  We also explore various Parameter Efficient Fine-Tuning (PEFT) methods to evaluate their practicality within the context of  information retrieval, shedding light on the effectiveness of different PEFT methods across diverse configurations with pre-training and model architectures influencing the choice of PEFT method.
 \\ \newline  }

\begin{document}

\maketitleabstract

\section{Introduction}
\input{text/introduction}

\section{Related Work}
\input{text/related}

\section{Task \& Dataset}
\input{text/dataset}

\section{Retrieval Models}
\input{text/models}

\section{Zero-shot \& Fine-tune Experiments}
\input{text/experiments-ret}

\section{Parameter-Efficient Retrieval}
\input{text/peft}

\section{PEFT Experiments}
\input{text/experiments-peft}

\section{Conclusion}
\input{text/conclusion}

\section{Limitations}
\input{text/limitations}

\section{Ethics Statement}
\input{text/ethics}

\section{Bibliographical References}\label{sec:reference}

\bibliographystyle{lrec-coling2024-natbib}
\bibliography{lrec-coling2024-example}



\end{document}

%% file: text/introduction.tex
Legal professionals including lawyers, judges and paralegals, often need to sift through voluminous legal judgments that encompass crucial insights for case law interpretations and judicial reasoning. These judgments, often lengthy, contain nuanced paragraphs holding the key to understanding legal principles, precedents and arguments. Finding relevant case law accounts for roughly 15 hours per week for a lawyer \cite{lastres2015rebooting} or nearly 30\% of their annual working hours \cite{poje2014legal}. Recent advances in NLP offer new possibilities to bridge this gap by providing summaries of these documents (e.g., \citealt{bhattacharya2019comparative,shukla2022legal} \textit{inter alia}). 
Nonetheless, practitioners still face challenges in navigating these texts to uncover specific paragraphs that address their queries. The current manual approach is labor-intensive and susceptible to overlooking essential details. Automating this process of identifying paragraphs relevant to the query streamlines legal research, allowing them to access relevant information efficiently.

Finding relevant paragraphs to a query is a challenging task unlike traditional adhoc information retrieval. Firstly, the legal domain is characterized by a vast and intricate vocabulary, interwoven with domain-specific jargon that can vary across different legal jurisdictions. This linguistic complexity demands an in-depth understanding of nuanced legal concepts, posing a substantial challenge for automated systems. The variation in legal writing style further compounds the challenge. Judgments may employ different degrees of formalism and offer varying levels of explicitness. These nuances can lead to difficulties in discerning context and accurately identifying relevant paragraphs that address specific queries. Another key challenge stems from the evolving nature of the legal case law. New legal doctrines, precedents and interpretations continually emerge, leading to an ever evolving array of legal concepts and principles. This dynamism necessitates a flexible and adaptive approach to comprehend new queries and determine relevance.  

To investigate the ability of current retrieval models to identify relevant paragraphs, a high-quality labeled dataset is imperative. However, creating such datasets is resource-intensive, often necessitating the involvement of legal experts to produce queries and  relevance labels. In this study, we employ distant supervision to construct a dataset tailored for the task of query-driven relevant paragraph extraction from legal judgments by the European Court of Human Rights (ECtHR) which addresses grievances by individuals against states for alleged violations of rights outlined in the European Convention of Human Rights. Our approach capitalizes on the case-law guides available through the ECtHR's Knowledge Sharing platform\footnote{\url{https://www.echr.coe.int/knowledge-sharing}}. We pose the case-law guide's section headers as queries, mirroring the legal concepts professionals utilize when searching within ECtHR judgments. We gather relevance signals by identifying the pinpointed citations to the paragraphs in the judgments within these guides under each section. 
Further, we meticulously design various splits to assess the generalizability of systems towards new queries (legal concepts), adapting to the evolution of law.\footnote{Our dataset is made available at \url{https://github.com/TUMLegalTech/ParagraphRetrievalECHR/}}

As a second contribution, we assess the performance of current retrieval models in a zero-shot manner using our dataset and further establish fine-tuning benchmarks employing diverse retrieval techniques encompassing dense bi-encoder and cross-encoder architectures. Our experiments reveal the drastic gap between fine-tuned and the zero-shot performance. Furthermore, we investigate into the efficacy of fine-tuning a general pre-trained model that was fine-tuned using other retrieval datasets (such as BERT fine-tuned on MSMARCO), comparing it against a legally pre-trained model (such as LegalBERT) that remains untouched by other retrieval datasets except ours. This investigation revealed that legal pre-training helps to handle distribution shift of the corpus, but still lacks in handling the distribution shift towards unseen queries. 

While complete fine-tuning has shown better performance, the trend towards larger models with billions or trillions of trainable parameters makes this fine-tuning process resource-intensive and costly. This spurred the exploration of Parameter Efficient Fine-Tuning (PEFT) strategies which update only a small number of extra parameters while keeping the original pre-trained model parameters frozen. In our study, we delve into this emerging area by evaluating representative methods of PEFT, namely Adapter \cite{houlsby2019parameter}, prefix-tuning \cite{li2021prefix} and LoRA \cite{hu2021lora}, within the context of our paragraph retrieval dataset.  This investigation contributes to the ongoing discourse regarding the practicality of adopting PEFT in the realm of Information Retrieval \cite{pal2023parameter,tam2022parameter,ma2022scattered,jung2022semi}. Our experiments demonstrate that PEFT methods achieve comparable performance to full fine-tuning on both seen and unseen queries, with the choice of the best PEFT method contingent on configuration such as general vs. legal pre-training and bi- vs. cross-encoder settings.

%% file: text/related.tex
\noindent \textbf{Legal IR}  Retrieving essential legal information is integral to the workflow of lawyers, encompassing tasks such as searching for legislation (adhoc search or by providing a factual description to identify the relevant statutes \cite{wang2018modeling,paul2022lesicin}), similar prior cases \cite{rabelo2022overview,mandal2017overview}, civil codes \cite{kim2016legal,kim2014answering}, litigation documents such as technology-assisted-review \cite{cormack2010overview}, patents \cite{piroi2013overview} and within law firm's internal support system \cite{moens2001innovative}. Our work focuses specifically on legal case retrieval. Most of the existing legal case law retrieval works primarily aim to retrieve entire cases \cite{sansone2022legal} based on different query granularities, including whole cases \cite{rabelo2022overview,ma2021lecard,mandal2017overview} or specific legal queries \cite{locke2017automatic,locke2018test,koniaris2016multi}. In contrast, our approach involves retrieving relevant paragraphs at a finer granularity, providing practitioners with a more targeted means of identifying essential information. At the paragraph granularity level, the legal case entailment task in COLIEE involves identifying a paragraph from existing cases that matches the decision of a new case \cite{rabelo2022overview}, but it employs the entire case as the query, in contrast to the short queries used in our work. This paragraph-level retrieval functionality is integral to building legal Question Answering \cite{khazaeli2021free, verma2020relevant} and Query-focused summarization systems.
\newline 

\noindent \textbf{Tasks on ECtHR Corpora} Previous works involving ECtHR corpus has dealt with judgement prediction \cite{aletras2016predicting,chalkidis2019neural,chalkidis2021paragraph,santosh2022deconfounding,tyss2023zero,tyss2023leveraging,xu2023dissonance}, argument mining \cite{mochales2008study,habernal2023mining,poudyal2019using,poudyal2020echr}, vulnerability detection \cite{xu2023vechr}, event extraction \cite{filtz2020events,navas2022whenthefact}. In this work, we capitalize on the case law guides maintained by registry of ECtHR to derive a query-driven relevant paragraph extraction dataset. We offer this dataset to the research community to facilitate  advancements 
in area of AI-enabled tools for legal practitioners.
\newline 

\noindent \textbf{Parameter Efficient Retrieval} With sizes of pre-trained language models  soaring up \cite{brown2020language}, full-parameter fine-tuning has become more challenging, this has created an interest in PEFT methods such as prompt tuning \cite{li2021prefix,lester2021power,liu2022p}, adapters  \cite{houlsby2019parameter,pfeiffer2021adapterfusion,mahabadi2021parameter}, additive methods \cite{hu2021lora,guo2021parameter,zhang2020side} and hybrid methods \cite{mao2022unipelt,chen2022parameter}. Specifically in IR, \citealt{ma2022scattered} conducted a comprehensive study of several PEFT methods for both the retrieval and re-ranking stages. \citealt{jung2022semi} has explored prefix-tuning and LoRA on bi-encoder models. \citealt{tam2022parameter} examined the effect of these methods on in-domain, cross-domain and cross-topic retrieval. \citealt{pal2023parameter} studied the effect of adapters on sparse retrieval models  contrary to dense models. We contribute to this ongoing discourse using both bi- and cross-encoders using our paragraph retrieval dataset on legal judgements.

%% file: text/dataset.tex
Our task of query-driven relevant paragraph extraction from legal judgements is defined as follows: Given a query $Q$ and a judgement document $J$ composed of $n$ paragraphs $P_J = \{p_1, p_2, \ldots, p_n\}$, the objective is to identify the subset of paragraphs $P^+_J \in P_J$ which are relevant to the query.

\subsection{Dataset Creation}
\textbf{Judgements Collection} We acquire ECtHR judgements collection as an HTML data dump from HUDOC \footnote{\url{http://hudoc.echr.coe.int/}}, the publicly available database of the ECtHR, along with their associated metadata. We retain only the English documents based on their metadata (Document Type: `HEJUD'). The parsing of judgment into paragraphs posed challenges due to inconsistent HTML structure, the presence of sub-paragraph numbers within each paragraph and the occurrence of spurious paragraph numbers resulting from verbatim text copied from other documents to cross-reference those paragraphs. To address these issues, we devised a range of hand-crafted heuristics to segment the judgment documents into paragraphs. Each paragraph is uniquely identified by its paragraph number at the beginning, facilitating cross-referencing.
\newline 

\noindent \textbf{Queries and Paragraph Relevance Collection} We curate our query-paragraph relevance dataset using case-law guides accessible on ECtHR Knowledge Sharing Platform\footnote{\url{https://www.echr.coe.int/knowledge-sharing}}. This platform, maintained by the court's registry, analyzes case law development for each convention article (e.g., Article 4 - Prohibition of slavery and forced labor\footnote{\url{https://ks.echr.coe.int/web/echr-ks/article-4}}) and transversal themes (e.g., Data Protection\footnote{\url{https://ks.echr.coe.int/web/echr-ks/data-protection}}, Rights of LGBTI persons\footnote{\url{https://ks.echr.coe.int/web/echr-ks/rights-of-lgbti-persons}}). It comprises 28 article and 8 theme-related case law guides, updated weekly, making them up-to-date with evolving case law with every new judgement and our proposed task  can in turn assist registry in achieving this goal of updating these guides regularly.
\newline 

\begin{figure}[!ht]
    \centering
    \includegraphics[width=0.85\linewidth]{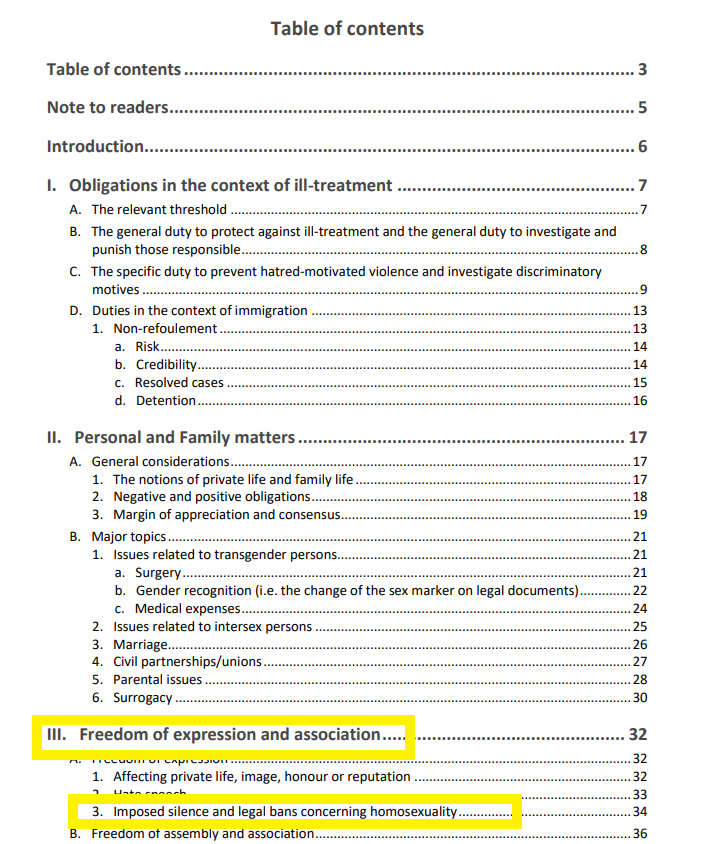}
    \caption{Query construction process from case law guide. The above table of contents is obtained from `Rights of LGBTI persons' guide.}
    \label{index-lgbti}
\end{figure}

\noindent \textbf{Obtaining queries} The case law guides provide the details of the key concepts involved under each article/theme and discuss them in detail by providing references to the relevant judgements. The legal concepts involved under each article/theme are structured in a hierarchical fashion, with sub-concepts enumerated.  A representative index structure of a case law guide is illustrated in Figure \ref{index-lgbti}. For instance, this is a hierarchical path of sections within the theme guide of Rights of LGBTI persons $\to$ Freedom of expression and association $\to$ Imposed silence and legal bans concerning homosexuality. We can extract this hierarchical structure by parsing the PDF case law guides' structural information. To construct each query, we combine these multiple concepts along the path (from the article or theme title to the leaf node in the PDF structure) by using a delimiter. This approach generates queries that mirror lists of legal concepts, akin to those sought after by legal practitioners when searching in ECtHR judgments. These queries/legal concepts could be used to index legal analytics databases that inform litigation strategies.
\newline 

\begin{figure}[!ht]
    \centering
    \includegraphics[width=0.85\linewidth]{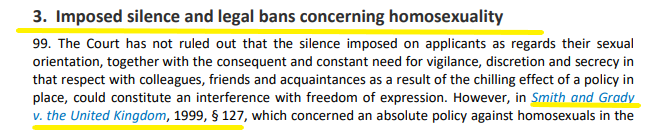}
    \caption{Illustration of pin-pointed paragraph relevance in case law guides.}
    \label{citation}
\end{figure}

\noindent \textbf{Obtaining relevant paragraphs in Judgements} These case-law guides provide in-depth discussions of each legal concept, offering pin-pointed paragraph references to the judgements from the ECtHR. An example of a legal concept description from a case-law guide is depicted in Fig.  \ref{citation}, demonstrating how relevant paragraphs are referenced under each query.  
We gather all paragraph references in a specific judgement under each legal concept and mark all of them as relevant corresponding to the given query in that judgement. However, it's worth mentioning that all judgements are not exhaustively covered in the case-law guide unless they contribute to the expansion or contraction of existing case law. Taking this into account, we pair queries with specific judgements referenced within them, subsequently extracting relevant paragraphs from these judgements. This contrasts with using all the paragraphs from all the judgements as the candidate set for identifying relevance. While our proposed methodology could theoretically be applied to all judgements across the corpus, we opt to restrict each query to the judgements specifically referenced under it. This deliberate limitation aims to ensure a high-quality evaluation setup, controlling false negatives. 

We filter out those query-judgement pairs in which reference to judgement is missing paragraph-level reference. Finally, we map back judgements in query-judgement pairs to our judgements collection, removing the ones  which we could not map back as some may refer to non-English documents which have not considered in our collection. 

\subsection{Data Splits \& Analysis}
\begin{figure*}[!ht]
    \centering
    \begin{subfigure}{0.22\textwidth}
       \includegraphics[width=1\linewidth]{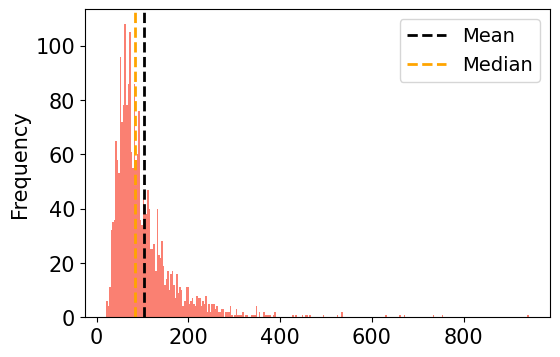}
    \caption{Number of paragraphs per judgement.}
    \label{num-para-dist}
    \end{subfigure}
    \hfill 
    \begin{subfigure}{0.22\textwidth}
       \includegraphics[width=1\linewidth]{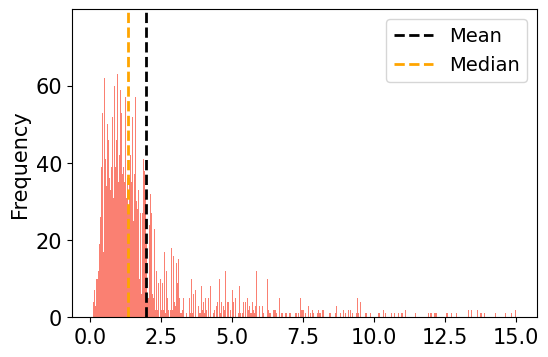}
    \caption{\% of relevant paragraphs per judgement.}
    \label{rel-para-dist}
    \end{subfigure}
    \hfill
    \begin{subfigure}{0.22\textwidth}
        \includegraphics[width=1\linewidth]{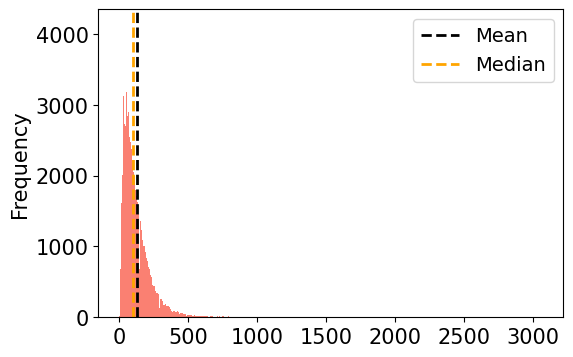}
    \caption{Number of tokens per paragraphs.}
    \label{para-len}
    \end{subfigure}
    \hfill
    \begin{subfigure}{0.22\textwidth}
        \includegraphics[width=1\linewidth]{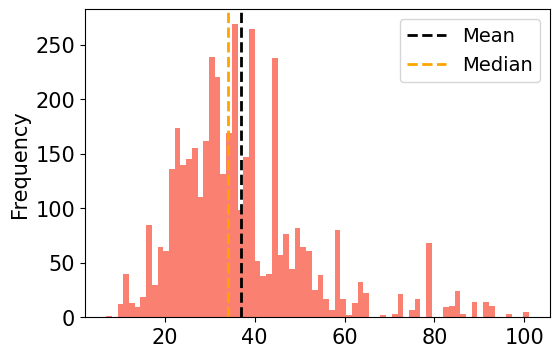}
    \caption{Number of tokens per query.}
    \label{query-len}
    \end{subfigure}
    \caption{Data Analysis}
\end{figure*}

We eventually end up with 4109 query-judgement pairs with 708 unique queries. The number of paragraphs in Judgement range from 21 to 942 with a mean of 102.78 (Fig. \ref{num-para-dist})
. The percentage of relevant paragraphs in each query-judgement pair range from 0.10\% to 15\% to the total number of paragraphs in that judgement with a mean around 1.95\%, depicted in Fig. \ref{rel-para-dist}. The queries and paragraph have a mean length of 36 and 135 tokens,  illustrated in Figures \ref{para-len} and \ref{query-len} respectively. 

We partition the article/theme case law guides into two distinct splits: one exclusively designated for testing with 403 query-judgment pairs (111 unique queries) derived from these case law guides, referred to as `Unseen article/themes'. This  creates a rigorous unseen evaluation scenario, assessing the model's performance on unfamiliar legal concepts from themes and articles that were not encountered during training. Queries originating from the other split are further divided into two subsets, resulting in `Seen article/theme, Unseen Query' with 694 pairs (120 unique queries) and `Seen article/theme, Seen Query' with 3012 pairs (477 unique queries). The former, reserved for testing, exposes the model to previously encountered themes/articles, but with new queries. The latter group is further divided into training (2230 pairs), validation (302 pairs), and test (480 pairs) sets. The test set within the `Seen article/theme, Seen Query' category assesses the model's comprehension of familiar legal concepts on new judgments in the test set. 

%% file: text/models.tex
We benchmark our task of identifying relevant paragraphs from a legal judgement given a query using the following models. We compute relevance score for each paragraph in given judgement with respect to the query and obtain the top-k most relevant paragraphs with the highest scores.
\newline

\noindent \textbf{BM25} \cite{robertson1995okapi} is a bag-of- words approach that estimates paragraph relevance to a query by considering the presence of query terms in the paragraph.
\newline 

\noindent \textbf{Bi-encoders}
employ separate encoders to encode queries and paragraphs into low-dimensional representations independently, leveraging neural architectures to capture semantic relationship and the final relevance score is computed using dot-product between the representations of query and paragraph obtained from encoder as $rel(q,p) = E_q(q).E_p(p)$ where $E_q$ and $E_p$ represent query and paragraph encoder respectively. The training objective 
is to learn representations such that relevant pairs of query and paragraphs will have higher similarity than the irrelevant ones. To reduce the training cost given there are lot of irrelevant paragraphs, negative sampling has been employed.
Let $\{⟨q_i, p^+_i , p^-_{i,1}, \ldots p^-_{i,n}⟩\}_{i=1}^m$ be the training data that consists of m instances with each instance consisting of one query $q_i$ and one relevant passage $p^+_i$, along with n irrelevant (negative) passages $p^-_{i,j}$. Note these negative paragraphs for a query are sampled from the same document as positive. We optimize negative log likelihood loss function as:
\begin{equation}
\scalemath{0.8}{
    L = -log(\frac{\exp(rel(q_i,p^+_i))}{\exp(rel(q,p^+_i)) + \sum_{j=1}^n \exp(rel(q,p^-_{i,j}))})
    }
\end{equation}
Following \citealt{karpukhin2020dense}, we consider negatives chosen from the irrelevant paragraphs randomly and the top paragraphs returned by BM25 which are not relevant to the query. We refer this approach as Dense Passage Retrieval \textbf{(DPR)}.  

Recently, \citealt{xiong2020approximate} proposed Approximate nearest neighbor Negative Contrastive Learning (\textbf{ANCE}) mechanism for dense retrieval. Instead of random or static BM25 negatives, ANCE constructs negatives using the being-optimized dense retrieval model. This helps to align the distribution of negative samples based on the models' training dynamics. While the model undergoes updates with each iteration, it would be expensive to update the negatives for every batch based on the updated model. Hence we asynchronously refresh the negatives at every checkpoint to reduce the computational cost to construct them.

These above methods follow a single-vector paradigm where each query and each paragraph is encoded into a single high-dimensional vector which is used to calculate relevance using a dot product. \citealt{khattab2020colbert} proposed a late interaction method named contextualized late interaction over BERT (\textbf{ColBERT}) where queries and documents are encoded at a finer granularity into multi-vector representations and relevance is estimated using interactions between these two sets of vectors. ColBERT produces an embedding for every token in the query and the paragraph and computes relevance as the sum of maximum similarities between each query vector and all vectors in the document as $rel(q,p) =  \sum_{i=1}^N \max_{j=1}^M Q_i.D_j^\intercal$  where Q is a query encoding matrix corresponding to N token vectors and D denotes the paragraph encoding matrix corresponding to M token vectors.
\newline 

\noindent \textbf{Cross-encoders} concatenate both of them before being provided to the model instead of encoding query and paragraph separately. The relevance score is directly computed by feed-forward network using the combined representation of the both \cite{yates2021pretrained} as $ rel(q,p) =  f(E_\phi(q,p))$
where $E_\phi$ represents a pre-trained model such as BERT and $f$ denotes a feed-forward network which takes [CLS] representation as input to compute relevance score and is trained end-to-end with binary cross entropy loss. This allows for deeper interaction between the query and paragraph but this effectiveness comes with a cost on efficiency as it now involves whole pass through the model for each query paragraph pair, instead of being able to pre-compute all the paragraph representations and use the model once to obtain query representation to calculate the relevance score as in bi-encoders.
   


%% file: text/experiments-ret.tex
Initially, we investigate the performance of retrieval models in a zero-shot evaluation scenario, where models trained on the MS MARCO paragraph ranking dataset \cite{bajaj2016ms} - a large-scale adhoc retrieval dataset derived from the Bing search log containing 8.8 million passages and around 800K queries for training, are directly evaluated on our legal judgement paragraph ranking dataset. We examine the following models: (i) DPR\footnote{\url{https://huggingface.co/facebook/dpr-question\_encoder-multiset-base}} (ii) ANCE\footnote{\url{https://huggingface.co/sentence-transformers/msmarco-roberta-base-ance-firstp}} (iii) ColBERT \footnote{\url{https://github.com/stanford-futuredata/ColBERT}} (iv) Cross encoder\footnote{\url{https://huggingface.co/cross-encoder/ms-marco-MiniLM-L-12-v2}}. We also evaluate a legal-domain-specific encoder model, LegalBERT \cite{chalkidis2020legal} which is pre-trained on diverse English legal texts encompassing legislative content, court cases, and contracts using cosine similarity between obtained [CLS] embeddings as relevance score. Notably, LegalBERT has been exposed to case law from ECtHR. 

Subsequently, we fine-tune these models on the training split of our legal judgment paragraph extraction dataset. We create two variants of each model, with distinct initializations: (i) model already fine-tuned on MSMARCO (models used in the zero-shot evaluation) and (ii) LegalBERT. 
\newline

\noindent \textbf{Implementation Details}
\label{ret-imp}
For DPR, we use mix of negatives from BM25 and random in ratio of 4:1 and train with total of 5 negatives per query-positive pair. For ANCE, we use same number of negatives derived from model. While for COLBERT and cross encoders, we use seven negatives samples for every positive query, where 4 are sampled randomly and 3 are from BM25 negatives. We sweep over learning rates $\{1e-5, 3e-5, 5e-5, 1e-4, 3e-4\}$ and the model is trained end-to-end for 5 epochs with Adam optimizer \cite{kingma2014adam} and we select the best model based on the performance on the validation set. 
\newline 

\noindent \textbf{Metrics} We evaluate the  performance using Recall@k\% (R@K\%). Recall@k\% measures the proportion of relevant paragraphs in the top-k\% of the total  paragraphs in the judgement and we report mean across all instances. We report for $k=\{2,5,10\}$. We use the k as percentage instead of absolute value to account for varying number of paragraphs across different judgements. Higher recall scores indicate better performance. 
\input{text/tab-ret}

\subsection{Results}
We report the results of both the zero-shot and the fine-tuning experiments in Table \ref{tab-ret}.

\noindent \textbf{Zero-shot:} We observe neural models demonstrate better performance across all the splits compared to BM25, bridging the lexical gap issue. ANCE displays slightly better performance than DPR demonstrating effectiveness of its dynamic negative sampling. COLBERT demonstrates superior performance across all variants, with a larger margin. This can be owed to its multi-vector representations at the granularity of each token and its training with  distillation loss from re-ranker models. We notice cross encoder are comparable to other dense models except to COLBERT, due to its ability to act better in re-ranking stage rather than retrieval stage. The performance order of these models is consistent with the out-of-domain zero-shot results on BEIR leaderboard\footnote{\url{https://github.com/beir-cellar/beir/wiki/Leaderboard}} \cite{thakur2021beir}. Surprisingly, LegalBERT performs comparably similar or less than BM25 and significantly below retrieval fine-tuned general models, contrary to what one might expect that legal pre-training would mitigate distribution shift on the corpus side to capture relevance. This points out general masked language model objective can not effectively translate to capture relevance in retrieval settings and calls for investigation of pre-training objectives suitable for retrieval such as inverse cloze task \cite{lee2019latent}, masking salient spans \cite{singh2021end} to handle phrase level query matching and contrastive based pre-training \cite{izacard2022unsupervised}. 
\newline 

\noindent \textbf{Zero-shot vs Fine-tune:} All the fine-tuning models (both MSMARCO and LegalBERT initialized ones) substantially improve over zero-shot variants in all the three splits. This difference highlights the need for future research to improve the generalization ability of current IR models to domains without any relevance label by handling distribution shifts from both the query and corpus side.
\newline 

\noindent \textbf{Fine-tune:} Despite COLBERT demonstrating a better zero-shot performance, cross encoders performed better with fine-tuning due to their deep interactions through concatenations, but that comes at a cost of efficiency to compute joint representation. Among bi-encoders, COLBERT perform well compared to ANCE followed by DPR due to its late interaction using multiple vector representations. The difference between them gets closer with fine-tuning on the `Seen Article, Seen Query' split, adapting the model to those specific queries. Across the other splits, we notice fine-tuning in general brings improvement over the zero-shot. However, the difference of improvement decreases with `Seen Article, Unseen Query' setting which further decreases with `Unseen article' setting. This highlights the need of effective strategies for domain adaptation with minimal labeled domain data without getting overfitted to those specific seen queries and handle distribution shift on query side. 
\newline 

\noindent \textbf{MSMARCO vs Legal} Across all the four models, we observe LegalBERT initialization outperforms MSMARCO variant, despite the opposite trend in zero-shot performance. This is more noticeable in unseen splits, where the legal pre-training helps the model in grasping context from the under specified queries compared to general pre-trained model with exposure to general factual-based QA instances. To unveil this capability of LegalBERT in zero-shot setup, it is crucial to design a pre-training objectives closely related to the retrieval task, as discussed before, to address the task shift.

This meticulous design of three different splits, coupled with these results highlight that this dataset can serve as 
 a testbed to study how to adapt these IR models to the distribution shifts between the source training task (such as MS MARCO) and the target tasks (such as ours) in zero-shot setup and also with minimal labeled data with some specific queries.

%% file: text/tab-ret.tex
\begin{table*}[]
\centering
\scalebox{0.98}{
\begin{tabular}{|l|l|l|ccc|ccc|ccc|}
\hline
\multirow{2}{*}{\textbf{}}                                           & \multicolumn{2}{l|}{\multirow{2}{*}{\textbf{}}}              & \multicolumn{3}{c|}{\textbf{\begin{tabular}[c]{@{}c@{}}Seen Article\\ Seen Query\end{tabular}}} & \multicolumn{3}{c|}{\textbf{\begin{tabular}[c]{@{}c@{}}Seen Article\\ Unseen Query\end{tabular}}} & \multicolumn{3}{c|}{\textbf{Unseen Article}}                                          \\ \cline{4-12} 
& \multicolumn{2}{l|}{}                                        & {\textbf{2\%}}     & {\textbf{5\%}}    & \textbf{10\%}    & {\textbf{2\%}}     & {\textbf{5\%}}     & \textbf{10\%}     & {\textbf{2\%}} & {\textbf{5\%}} & \textbf{10\%} \\ \hline
\multirow{6}{*}{\begin{tabular}[c]{@{}l@{}}Zero\\ shot\end{tabular}} & \multicolumn{2}{l|}{BM25}                                    & {0.07}             & {0.17}            & 0.29             & {0.09}             & {0.23}             & 0.37              & {0.10}          & {0.25}         & 0.40         \\ \cline{2-12} 
& \multicolumn{2}{l|}{DPR}                                     & {0.11}             & {0.22}            & 0.33             & {0.14}             & {0.26}             & 0.42              & {0.14}         & {0.30}          & 0.47          \\ \cline{2-12} 
& \multicolumn{2}{l|}{ANCE}                                    & {0.12}             & {0.23}            & 0.34             & {0.16}             & {0.28}             & 0.44              & {0.17}         & {0.34}         & 0.48          \\ \cline{2-12} 
& \multicolumn{2}{l|}{COLBERT}                                 & {0.16}             & {0.32}            & 0.47             & {0.17}             & {0.34}             & 0.51              & {0.24}         & {0.41}         & 0.56          \\ \cline{2-12} 
& \multicolumn{2}{l|}{CrossEncoder}                            & {0.08}             & {0.20}             & 0.35             & {0.15}             & {0.28}             & 0.42              & {0.20}          & {0.36}         & 0.50         \\ \cline{2-12} 
& \multicolumn{2}{l|}{LegalBERT}                               & {0.06}             & {0.16}            & 0.32             & {0.09}             & {0.23}             & 0.37              & {0.08}         & {0.21}         & 0.36          \\ \hline
\multirow{8}{*}{\begin{tabular}[c]{@{}l@{}}Fine\\ tune\end{tabular}} & {\multirow{2}{*}{DPR}}          & MSMARCO & {0.21}             & {0.41}            & 0.60              & {0.22}             & {0.40}              & 0.60               & {0.25}         & {0.45}         & 0.64          \\ 
& {}                              & Legal   & {0.28}             & {0.47}            & 0.65             & {0.24}             & {0.46}             & 0.67              & {0.29}         & {0.50}          & 0.68          \\ \cline{2-12} 
& {\multirow{2}{*}{ANCE}}         & MSMARCO & {0.22}             & {0.43}            & 0.62             & {0.24}             & {0.41}             & 0.61              & {0.26}         & {0.46}         & 0.66          \\ 
& {}                              & Legal   & {0.28}             & {0.48}            & 0.67             & {0.24}             & {0.47}             & 0.68              & {0.26}         & {0.51}         & 0.69          \\ \cline{2-12}
& {\multirow{2}{*}{COLBERT}}      & MSMARCO & {0.25}             & {0.45}            & 0.64             & {0.27}             & {0.46}             & 0.66              & {0.25}         & {0.49}         & 0.69          \\ 
& {}                              & Legal   & {0.29}             & {0.49}            & 0.69             & {0.29}             & {0.49}             & 0.69              & {0.27}         & {0.51}         & 0.70           \\ \cline{2-12} 
&  {\multirow{2}{*}{\begin{tabular}[c]{@{}l@{}}Cross\\ Encoder\end{tabular}}} & MSMARCO & {0.26}             & {0.48}            & 0.69             & {0.30}              & {0.50}              & 0.71              & {0.31}         & {0.51}         & 0.70           \\ 
& {}                              & Legal   & {0.30}              & {0.50}             & 0.70              & {0.31}             & {0.54}             & 0.72              & {0.32}         & {0.57}         & 0.74          \\ \hline
\end{tabular}}
\caption{Results of various systems on our Query-driven Paragraph retrieval task. For zero-shot settings, all these splits are unseen, as they are not fine-tuned on any task related data.}
\label{tab-ret}
\end{table*}

%% file: text/peft.tex
PEFT aims to tune only a small portion of parameters rather than the full parameters as in traditional fine-tuning. PEFT approaches fall into three primary categories: Parameter Composition, Input Composition, and Function Composition \cite{ruder2022modular}. Given a neural network $f_\theta: X \to Y$, it is decomposed into a sequence of functions $f_\theta = f_{\theta_1} \odot f_{\theta_2} \odot \ldots f_{\theta_l}$, where $\theta_1, \theta_2, \ldots, \theta_l$ represent parameters which are held constant in PEFT and a module with parameters $\phi$ is introduced, which are updated during training to modify the $i^{th}$ sub-function as follows:
Parameter composition involves interpolating models' parameter with new parameters as $f'_i(x) = f_{\theta_i \odot \phi}(x)$. Input Composition augments a model’s input with a learnable parameter vector as $f'_i(x) = f_{\theta_i}([x,\phi]) $
Function composition augments a model’s functions with new task-specific functions as $f'_i(x) = f_{\theta_i} \odot f_{\phi}(x) $. We pick one representative method from each category and study their performance on our retrieval task.
\newline

\noindent \textbf{Adapters} \cite{houlsby2019parameter} fall under the category of function composition where we inject two small modules between the self-attention sub-layer and the feed forward sub-layer inside each layer of transformer sequentially. The adapter module consists of a down-projection, an up-projection and a nonlinear function between them with a residual connection across each module.
\begin{equation}
    Adapter(h) = h + W_{up}^\intercal \psi(W_{down}^\intercal h) 
\end{equation}
where $W_{down} \in \mathbb{R}^{D_{hidden}\times D_{mid}}$ and $W_{up} \in \mathbb{R}^{D_{mid} \times D_{hidden}}$, $D_{mid}$ denote the bottleneck dimension and $\psi$ is a nonlinear RELU activation function.
\newline 

\noindent \textbf{Prefix-Tuning} \cite{li2021prefix}  falls under the category of input-composition where we prepend a fixed number of trainable vectors to the input of multi-head attention in each Transformer layer, which the original tokens can attend to as if they were virtual tokens. Specifically two prefix matrices $P_K$ and $P_V \in \mathbb{R}^{L \times D_{hidden}}$ are prepended to K and V where L denotes prefix length. 
\begin{equation}
h = Attention(Q, [P_k, k], [P_v,v])
\end{equation}
\noindent \textbf{LoRA} \cite{hu2021lora} Low-Rank Adaptation falls under the category of parameter-composition, introduces trainable low-rank matrices and combines them with the original matrices in the multi-head attention. Specifically, it learns two low-rank matrices $W_{down} \in \mathbb{R}^{D_{hidden} \times D_{mid}}$ and $W_{up} \in \mathbb{R}^{D_{mid}\times D_{hidden}}$ for each of the query and value projections along with their original matrix $W_Q$ and $W_V \in \mathbb{R}^{D_{hidden} \times D_{hidden}}$. Taking $W_Q$ as example:
\begin{equation}
    Q = (W_Q^\intercal + \alpha W_{up}^\intercal W_{down}^\intercal) h_{in}
\end{equation}
where $\alpha$ is a tunable hyper-parameter. Once after the training is complete, we can sum up these additional LoRA weights to the original weights, thus making the inference overhead to zero. 
\newline

%% file: text/experiments-peft.tex
We investigate the effect of PEFT by applying each method separately on bi-encoder and cross-encoder, using MSMARCO and LegalBERT initializations. Among bi-encoders, we choose COLBERT due to its better performance in full fine-training. 
We report  Recall@k\% for $k=\{2,5,10\}$ in Table \ref{peft-tab}.

\input{text/tab-peft}

\noindent \textbf{Implementation Details}
\label{peft-imp}
We use the AdapterHub library\footnote{\url{https://docs.adapterhub.ml}} for implementing PEFT methods. For Prefix-tuning, we use prefix lengths of 10, 15 and 30. For Bottleneck adapters, we used reduction factors of 8, 16, and 32. In case of LoRA, we use configuration of rank  and alpha in $\{8, 16\}$. We sweep over learning rates $\{1e-5, 3e-5, 5e-5, 1e-4, 3e-4\}$  select the best model based on the performance on the validation set. We train the model for 15 epochs with Adam optimizer \cite{kingma2014adam}.

\subsection{Results}
\noindent \textbf{CrossEncoder (MSMARCO):} We observe all the PEFT methods under perform than full fine-tuning across all the splits. Among them, LORA underperforms consistently across all the splits, while prefix tuning is better among them.  However, Adapter takes the lead in `unseen article' split and this can be attributed to better generalization capability derived through adding new functional composition rather than additional input tokens in case of prefix tuning, which may to overfit on seen article splits. 
\newline 

\noindent \textbf{CrossEncoder (Legal):} We observe all the PEFT methods comparable to each other across all the splits, which can be attributed to domain-specific legal knowledge from base model. On the `seen query' split, they even surpass the full fine-tuning, demonstrating that with fine-tuning $\sim$1\% of the original model parameters, they can achieve comparable performance to the full fine-tuning baseline, makes them to adopt easily in low-compute settings. However, these methods fall back on generalizability, compared to full-tuning, opening up potential directions to tackle in future, how to augment these PEFT methods to handle these distribution shifts to perform effectively on unseen settings.
\newline 

\noindent \textbf{COLBERT (MSMARCO):} PEFT methods underperform compared to full fine-tuning. Among them, Prefix Tuning turns out to be lowest performer and rest of them are comparable to each other, across all the splits consistently. This can be attributed to the short queries in our case. COLBERT (bi-encoder) models encode queries and paragraphs separately, and for shorter queries, they struggle to extract meaningful contextual information using BERT alone. Prefix Tuning, in particular, fails to enhance this contextual information just by adding additional parameters in the input compared to others which can handle the representation embeddings through function or parameter composition. 
\newline 

\noindent \textbf{COLBERT (Legal):} We observe similar to the COLBERT(MSMACRO), prefix tuning underperforming compared to rest of the methods. 
\newline 

\noindent \textbf{Cross encoder vs COLBERT:} While prefix tuning turned out to be a better PEFT method in cross encoder setting (especially in MSMARCO), it turned out to be lowest in bi-encoder, COLBERT, encouraging further studies to develop model agnostic PEFT methods and analyze the interplay between architecture and the PEFT method.  
\newline 

\noindent \textbf{MSMARCO vs Legal:} Overall, legal pre-training helped to account for distribution shift for corpus, demonstrating better results. This coupled with cross-encoder deep interactions, demonstrated parameter efficiency when fine-tuning. Moreover, Legal oriented models witness only a small decline with sparse fine-tuning from full fine-tuning in comparison to MSMARCO variants.

Overall, we empirically demonstrate that PEFT methods can achieve comparable performance  to full-parameter fine-tuning not only in seen query setting but also in challenging unseen settings and motivate further work to bridge the existing gap between them, making them more adaptable in low data and compute resource settings.

%% file: text/tab-peft.tex
\begin{table*}[]
\centering
\scalebox{0.95}{
\begin{tabular}{|l|l|c|ccc|ccc|ccc|}
\hline
\textbf{}                                                                          & \textbf{}    & \multirow{2}{*}{\textbf{\begin{tabular}[c]{@{}c@{}}\%\\ train\end{tabular}}}       & \multicolumn{3}{c|}{\textbf{\begin{tabular}[c]{@{}c@{}}Seen Article\\ Seen Query\end{tabular}}} & \multicolumn{3}{c|}{\textbf{\begin{tabular}[c]{@{}c@{}}Seen Article\\ Unseen Query\end{tabular}}} & \multicolumn{3}{c|}{\textbf{Unseen Article}}                                    \\ \cline{1-2} \cline{4-12} 
\textbf{}                                                                          & \textbf{}    &  & {\textbf{2\%}}       & {\textbf{5\%}}      & \textbf{10\%}      & {\textbf{2\%}}       & {\textbf{5\%}}       & \textbf{10\%}       & {\textbf{2\%}} & {\textbf{5\%}} & \textbf{10\%} \\ \hline
\multirow{4}{*}{\begin{tabular}[c]{@{}l@{}}Cross\\ Encoder\\ MSMARCO\end{tabular}} & Full      & 100                                                                          & {0.26}             & {0.48}            & 0.69             & {0.30}              & {0.50}              & 0.71              & {0.31}       & {0.51}       & 0.70        \\
& Adapter      & 1.6                                                                          & {0.25}             & {0.45}            & 0.63             & {0.28}             & {0.47}             & 0.67              & {0.30}        & {0.50}        & 0.68        \\ 
& Pre. Tun. & 0.5                                                                         & {0.27}             & {0.48}            & 0.65             & {0.31}             & {0.51}             & 0.69              & {0.28}       & {0.47}       & 0.66        \\  
& LORA         & 0.5                                                                          & {0.24}             & {0.42}            & 0.60             & {0.26}             & {0.45}             & 0.64              & {0.26}       & {0.46}       & 0.63        \\ \hline
\multirow{4}{*}{\begin{tabular}[c]{@{}l@{}}Cross \\ Encoder\\ Legal\end{tabular}}  & Full       & 100                                                                          & {0.30}              & {0.50}             & 0.70             & {0.31}             & {0.54}             & 0.72              & {0.32}       & {0.57}       & 0.74        \\ 
& Adapter      & 1.3                                                                         & {0.30}              & {0.52}            & 0.71             & {0.28}             & {0.49}             & 0.68              & {0.26}       & {0.48}       & 0.70        \\  
& Pre. Tun. & 0.8                                                                         & {0.30}              & {0.52}            & 0.71             & {0.29}             & {0.48}             & 0.68              & {0.27}       & {0.49}       & 0.70        \\ 
& LORA         & 0.9                                                                         & {0.29}             & {0.51}            & 0.70             & {0.28}             & {0.48}             & 0.69              & {0.27}       & {0.49}       & 0.70        \\ \hline
\multirow{4}{*}{\begin{tabular}[c]{@{}l@{}}COLBERT\\ MSMARCO\end{tabular}}         & Full       & 100                                                                          & {0.25}             & {0.45}            & 0.64             & {0.27}             & {0.46}             & 0.66              & {0.29}       & {0.49}       & 0.69        \\ 
 & Adapter      & 1.6                                                                          & {0.22}             & {0.41}            & 0.60             & {0.24}             & {0.43}             & 0.62              & {0.24}       & {0.43}       & 0.62        \\ 
& Pre. Tun. & 0.5                                                                         & {0.19}             & {0.39}            & 0.58             & {0.21}             & {0.40}              & 0.59              & {0.20}        & {0.39}       & 0.60        \\  
 & LORA         & 0.5                                                                          & {0.21}             & {0.41}            & 0.60             & {0.24}             & {0.42}             & 0.62              & {0.24}       & {0.43}       & 0.62        \\ \hline
\multirow{4}{*}{\begin{tabular}[c]{@{}l@{}}COLBERT\\ Legal\end{tabular}}           & Full       & 100                                                                          & {0.28}             & {0.48}            & 0.67             & {0.24}             & {0.47}             & 0.68              & {0.26}       & {0.51}       & 0.69        \\ 
& Adapter      & 1.6                                                                         & {0.26}             & {0.46}            & 0.64             & {0.25}             & {0.46}             & 0.67              & {0.23}       & {0.46}       & 0.64        \\ 
& Pre. Tun. & 0.5                                                                          & {0.20}              & {0.40}             & 0.61             & {0.21}             & {0.41}             & 0.61              & {0.19}       & {0.40}        & 0.57        \\ 
& LORA         & 0.5                                                                         & {0.26}             & {0.46}            & 0.63             & {0.24}             & {0.46}             & 0.66              & {0.24}       & {0.46}       & 0.63        \\ \hline
\end{tabular}}
\caption{Comparison between full fine-tuning and various parameter-efficient tuning methods.}
\label{peft-tab}
\end{table*}

%% file: text/conclusion.tex
We present an empirical study focused on the task of extracting relevant paragraphs from legal judgments based on the query. We rigorously curate a dataset for this task from ECtHR jurisdiction, leveraging the case-law guides produced by the court's registry.  We assess the current retrieval models on this task in a zero-shot way to emphasize the need of retrieval specific pre-training objectives. We further fine-tune several models encompassing bi- and cross-encoders for this task. We evaluate the generalizability of different fine-tuning models when faced with unseen concepts or queries to illustrate how legal pre-training can effectively address distribution shifts on the corpus side but still faces challenges in adapting to shift on the query side. In addition, we demonstrate the efficacy of different PEFT methods on these retrieval methods shedding light on their intricate effects concerning legal pre-training, bi-encoder, and cross-encoder models. Our findings reveal that there is no one-size-fits-all PEFT method that performs well across all settings. We hope that both our dataset and the fine-tuned models will be useful to the research community working in the space of legal information retrieval.

%% file: text/limitations.tex
In this study, we treat each paragraph as an independent unit during the training of neural models. However, it's important to note that paragraphs are not entirely independent; they often constitute small excerpts from longer documents, and their content may not always provide a comprehensive estimate of their relevance. To be more precise, some paragraphs draw context not just from other paragraphs within the same document but also from other documents, as evident through citations and cross-references to other judgments or texts within each paragraph. In the future, harnessing this inter-paragraph and cross-document contextual signal could lead to a more enriched understanding of each paragraph's relevance.

Furthermore, this practice of segmenting documents into smaller chunks is common in retrieval tasks, where documents are broken down into shorter lengths for indexing in retrieval systems. This may lead to losing some of their context in the process. It's worth noting that this chunking effect could be more pronounced in our task compared to more fact-focused general retrieval tasks. As a result, future research should explore methods to effectively capture contextual information from the sequential nature of paragraphs within documents and develop discourse-aware representations.

Most retriever systems follow a two-stage pipeline approach, where a pre-fetcher first aims to return all relevant ones followed by a re-ranker which attempts to make more relevant ones appear before less relevant ones. In this work, we explicitly focused our experiments on the pre-fetcher component leaving the second component for future work.

A specific challenge with respect to PEFT methods are that they converge slower and  relatively more sensitive to hyper-parameters such as  learning rate than full fine-tuning. We do observe same characteristics during our work  and have to bypass the problem by training for more epochs and experimenting with different hyper-parameters. It is thus imperative to analyse them more theoretically and design more robust and stable training strategies for PEFT methods in the future.

Additionally, while our findings of the relevant paragraph retrieval experiments are specific to the ECtHR domain and datasets, comparable experiments in other domains will see variation based on the nature of the legal concepts and the legal documents. Nevertheless, all derivation of insights from legal case data comes with jurisdiction-related limitations. 

%% file: text/ethics.tex
With the release of our data as a public resource, we do not foresee any ethical concerns in a repurposed and bundled release of this dataset as both the judgements data and the caselaw guides are already available through the HUDOC and ECHR KS platform respectively and it complies with the ECtHR data policy. These decisions, although not anonymized, include the real names of individuals involved. However, our work does not engage with the data in a way that we consider harmful beyond this availability. We believe that this work can foster further research in this data scarce legal NLP field, to build assistive technology for legal professionals. We are conscious by employing pre-trained language models, we inherit the biases they may have acquired from their training corpus and need to be further scrutinized of any biases that may arise and is crucial to ensure that the systems developed are fair.

%% file: lrec-coling2024-example.bbl
\begin{thebibliography}{63}
\expandafter\ifx\csname natexlab\endcsname\relax\def\natexlab#1{#1}\fi

\bibitem[{Aletras et~al.(2016)Aletras, Tsarapatsanis, Preo{\c{t}}iuc-Pietro,
  and Lampos}]{aletras2016predicting}
Nikolaos Aletras, Dimitrios Tsarapatsanis, Daniel Preo{\c{t}}iuc-Pietro, and
  Vasileios Lampos. 2016.
\newblock Predicting judicial decisions of the european court of human rights:
  A natural language processing perspective.
\newblock \emph{PeerJ computer science}, 2:e93.

\bibitem[{Bajaj et~al.(2016)Bajaj, Campos, Craswell, Deng, Gao, Liu, Majumder,
  McNamara, Mitra, Nguyen et~al.}]{bajaj2016ms}
Payal Bajaj, Daniel Campos, Nick Craswell, Li~Deng, Jianfeng Gao, Xiaodong Liu,
  Rangan Majumder, Andrew McNamara, Bhaskar Mitra, Tri Nguyen, et~al. 2016.
\newblock Ms marco: A human generated machine reading comprehension dataset.
\newblock \emph{arXiv e-prints}, pages arXiv--1611.

\bibitem[{Bhattacharya et~al.(2019)Bhattacharya, Hiware, Rajgaria, Pochhi,
  Ghosh, and Ghosh}]{bhattacharya2019comparative}
Paheli Bhattacharya, Kaustubh Hiware, Subham Rajgaria, Nilay Pochhi,
  Kripabandhu Ghosh, and Saptarshi Ghosh. 2019.
\newblock A comparative study of summarization algorithms applied to legal case
  judgments.
\newblock In \emph{Advances in Information Retrieval: 41st European Conference
  on IR Research, ECIR 2019, Cologne, Germany, April 14--18, 2019, Proceedings,
  Part I 41}, pages 413--428. Springer.

\bibitem[{Brown et~al.(2020)Brown, Mann, Ryder, Subbiah, Kaplan, Dhariwal,
  Neelakantan, Shyam, Sastry, Askell et~al.}]{brown2020language}
Tom Brown, Benjamin Mann, Nick Ryder, Melanie Subbiah, Jared~D Kaplan, Prafulla
  Dhariwal, Arvind Neelakantan, Pranav Shyam, Girish Sastry, Amanda Askell,
  et~al. 2020.
\newblock Language models are few-shot learners.
\newblock \emph{Advances in neural information processing systems},
  33:1877--1901.

\bibitem[{Chalkidis et~al.(2019)Chalkidis, Androutsopoulos, and
  Aletras}]{chalkidis2019neural}
Ilias Chalkidis, Ion Androutsopoulos, and Nikolaos Aletras. 2019.
\newblock Neural legal judgment prediction in english.
\newblock In \emph{Proceedings of the 57th Annual Meeting of the Association
  for Computational Linguistics}, pages 4317--4323.

\bibitem[{Chalkidis et~al.(2020)Chalkidis, Fergadiotis, Malakasiotis, Aletras,
  and Androutsopoulos}]{chalkidis2020legal}
Ilias Chalkidis, Manos Fergadiotis, Prodromos Malakasiotis, Nikolaos Aletras,
  and Ion Androutsopoulos. 2020.
\newblock Legal-bert: The muppets straight out of law school.
\newblock In \emph{Findings of the Association for Computational Linguistics:
  EMNLP 2020}, pages 2898--2904.

\bibitem[{Chalkidis et~al.(2021)Chalkidis, Fergadiotis, Tsarapatsanis, Aletras,
  Androutsopoulos, and Malakasiotis}]{chalkidis2021paragraph}
Ilias Chalkidis, Manos Fergadiotis, Dimitrios Tsarapatsanis, Nikolaos Aletras,
  Ion Androutsopoulos, and Prodromos Malakasiotis. 2021.
\newblock Paragraph-level rationale extraction through regularization: A case
  study on european court of human rights cases.
\newblock In \emph{Proceedings of the 2021 Conference of the North American
  Chapter of the Association for Computational Linguistics: Human Language
  Technologies}, pages 226--241.

\bibitem[{Chen et~al.(2022)Chen, Zhang, Shi, Li, Smola, and
  Yang}]{chen2022parameter}
Jiaao Chen, Aston Zhang, Xingjian Shi, Mu~Li, Alex Smola, and Diyi Yang. 2022.
\newblock Parameter-efficient fine-tuning design spaces.
\newblock In \emph{The Eleventh International Conference on Learning
  Representations}.

\bibitem[{Cormack et~al.(2010)Cormack, Grossman, Hedin, and
  Oard}]{cormack2010overview}
Gordon~V Cormack, Maura~R Grossman, Bruce Hedin, and Douglas~W Oard. 2010.
\newblock Overview of the trec 2010 legal track.
\newblock In \emph{TREC}.

\bibitem[{Filtz et~al.(2020)Filtz, Navas-Loro, Santos, Polleres, and
  Kirrane}]{filtz2020events}
Erwin Filtz, Mar{\'\i}a Navas-Loro, Cristiana Santos, Axel Polleres, and
  Sabrina Kirrane. 2020.
\newblock Events matter: Extraction of events from court decisions.
\newblock \emph{Legal Knowledge and Information Systems}, pages 33--42.

\bibitem[{Guo et~al.(2021)Guo, Rush, and Kim}]{guo2021parameter}
Demi Guo, Alexander~M Rush, and Yoon Kim. 2021.
\newblock Parameter-efficient transfer learning with diff pruning.
\newblock In \emph{Proceedings of the 59th Annual Meeting of the Association
  for Computational Linguistics and the 11th International Joint Conference on
  Natural Language Processing (Volume 1: Long Papers)}, pages 4884--4896.

\bibitem[{Habernal et~al.(2023)Habernal, Faber, Recchia, Bretthauer, Gurevych,
  Spiecker~genannt D{\"o}hmann, and Burchard}]{habernal2023mining}
Ivan Habernal, Daniel Faber, Nicola Recchia, Sebastian Bretthauer, Iryna
  Gurevych, Indra Spiecker~genannt D{\"o}hmann, and Christoph Burchard. 2023.
\newblock Mining legal arguments in court decisions.
\newblock \emph{Artificial Intelligence and Law}, pages 1--38.

\bibitem[{Houlsby et~al.(2019)Houlsby, Giurgiu, Jastrzebski, Morrone,
  De~Laroussilhe, Gesmundo, Attariyan, and Gelly}]{houlsby2019parameter}
Neil Houlsby, Andrei Giurgiu, Stanislaw Jastrzebski, Bruna Morrone, Quentin
  De~Laroussilhe, Andrea Gesmundo, Mona Attariyan, and Sylvain Gelly. 2019.
\newblock Parameter-efficient transfer learning for nlp.
\newblock In \emph{International Conference on Machine Learning}, pages
  2790--2799. PMLR.

\bibitem[{Hu et~al.(2021)Hu, Wallis, Allen-Zhu, Li, Wang, Wang, Chen
  et~al.}]{hu2021lora}
Edward~J Hu, Phillip Wallis, Zeyuan Allen-Zhu, Yuanzhi Li, Shean Wang, Lu~Wang,
  Weizhu Chen, et~al. 2021.
\newblock Lora: Low-rank adaptation of large language models.
\newblock In \emph{International Conference on Learning Representations}.

\bibitem[{Izacard et~al.(2022)Izacard, Caron, Hosseini, Riedel, Bojanowski,
  Joulin, and Grave}]{izacard2022unsupervised}
Gautier Izacard, Mathilde Caron, Lucas Hosseini, Sebastian Riedel, Piotr
  Bojanowski, Armand Joulin, and Edouard Grave. 2022.
\newblock Unsupervised dense information retrieval with contrastive learning.
\newblock \emph{Transactions on Machine Learning Research}.

\bibitem[{Jung et~al.(2022)Jung, Choi, and Rhee}]{jung2022semi}
Euna Jung, Jaekeol Choi, and Wonjong Rhee. 2022.
\newblock Semi-siamese bi-encoder neural ranking model using lightweight
  fine-tuning.
\newblock In \emph{Proceedings of the ACM Web Conference 2022}, pages 502--511.

\bibitem[{Karpukhin et~al.(2020)Karpukhin, Oguz, Min, Lewis, Wu, Edunov, Chen,
  and Yih}]{karpukhin2020dense}
Vladimir Karpukhin, Barlas Oguz, Sewon Min, Patrick Lewis, Ledell Wu, Sergey
  Edunov, Danqi Chen, and Wen-tau Yih. 2020.
\newblock Dense passage retrieval for open-domain question answering.
\newblock In \emph{Proceedings of the 2020 Conference on Empirical Methods in
  Natural Language Processing (EMNLP)}, pages 6769--6781.

\bibitem[{Khattab and Zaharia(2020)}]{khattab2020colbert}
Omar Khattab and Matei Zaharia. 2020.
\newblock Colbert: Efficient and effective passage search via contextualized
  late interaction over bert.
\newblock In \emph{Proceedings of the 43rd International ACM SIGIR conference
  on research and development in Information Retrieval}, pages 39--48.

\bibitem[{Khazaeli et~al.(2021)Khazaeli, Punuru, Morris, Sharma, Staub, Cole,
  Chiu-Webster, and Sakalley}]{khazaeli2021free}
Soha Khazaeli, Janardhana Punuru, Chad Morris, Sanjay Sharma, Bert Staub,
  Michael Cole, Sunny Chiu-Webster, and Dhruv Sakalley. 2021.
\newblock A free format legal question answering system.
\newblock In \emph{Proceedings of the Natural Legal Language Processing
  Workshop 2021}, pages 107--113.

\bibitem[{Kim et~al.(2014)Kim, Xu, Goebel, and Satoh}]{kim2014answering}
Mi-Young Kim, Ying Xu, Randy Goebel, and Ken Satoh. 2014.
\newblock Answering yes/no questions in legal bar exams.
\newblock In \emph{New Frontiers in Artificial Intelligence: JSAI-isAI 2013
  Workshops, LENLS, JURISIN, MiMI, AAA, and DDS, Kanagawa, Japan, October
  27--28, 2013, Revised Selected Papers 5}, pages 199--213. Springer.

\bibitem[{Kim et~al.(2016)Kim, Xu, Lu, and Goebel}]{kim2016legal}
Mi-Young Kim, Ying Xu, Yao Lu, and Randy Goebel. 2016.
\newblock Legal question answering using paraphrasing and entailment analysis.
\newblock In \emph{Tenth International Workshop on Juris-informatics
  (JURISIN)}.

\bibitem[{Kingma and Ba(2014)}]{kingma2014adam}
Diederik~P Kingma and Jimmy Ba. 2014.
\newblock Adam: A method for stochastic optimization.
\newblock \emph{arXiv preprint arXiv:1412.6980}.

\bibitem[{Koniaris et~al.(2016)Koniaris, Anagnostopoulos, and
  Vassiliou}]{koniaris2016multi}
Marios Koniaris, Ioannis Anagnostopoulos, and Yannis Vassiliou. 2016.
\newblock Multi-dimension diversification in legal information retrieval.
\newblock In \emph{Web Information Systems Engineering--WISE 2016: 17th
  International Conference, Shanghai, China, November 8-10, 2016, Proceedings,
  Part I 17}, pages 174--189. Springer.

\bibitem[{Lastres(2015)}]{lastres2015rebooting}
Steven~A Lastres. 2015.
\newblock Rebooting legal research in a digital age.

\bibitem[{Lee et~al.(2019)Lee, Chang, and Toutanova}]{lee2019latent}
Kenton Lee, Ming-Wei Chang, and Kristina Toutanova. 2019.
\newblock Latent retrieval for weakly supervised open domain question
  answering.
\newblock In \emph{Proceedings of the 57th Annual Meeting of the Association
  for Computational Linguistics}, pages 6086--6096.

\bibitem[{Lester et~al.(2021)Lester, Al-Rfou, and Constant}]{lester2021power}
Brian Lester, Rami Al-Rfou, and Noah Constant. 2021.
\newblock The power of scale for parameter-efficient prompt tuning.
\newblock In \emph{Proceedings of the 2021 Conference on Empirical Methods in
  Natural Language Processing}, pages 3045--3059.

\bibitem[{Li and Liang(2021)}]{li2021prefix}
Xiang~Lisa Li and Percy Liang. 2021.
\newblock Prefix-tuning: Optimizing continuous prompts for generation.
\newblock In \emph{Proceedings of the 59th Annual Meeting of the Association
  for Computational Linguistics and the 11th International Joint Conference on
  Natural Language Processing (Volume 1: Long Papers)}, pages 4582--4597.

\bibitem[{Liu et~al.(2022)Liu, Ji, Fu, Tam, Du, Yang, and Tang}]{liu2022p}
Xiao Liu, Kaixuan Ji, Yicheng Fu, Weng Tam, Zhengxiao Du, Zhilin Yang, and Jie
  Tang. 2022.
\newblock P-tuning: Prompt tuning can be comparable to fine-tuning across
  scales and tasks.
\newblock In \emph{Proceedings of the 60th Annual Meeting of the Association
  for Computational Linguistics (Volume 2: Short Papers)}, pages 61--68.

\bibitem[{Locke and Zuccon(2018)}]{locke2018test}
Daniel Locke and Guido Zuccon. 2018.
\newblock A test collection for evaluating legal case law search.
\newblock In \emph{The 41st International ACM SIGIR Conference on Research \&
  Development in Information Retrieval}, pages 1261--1264.

\bibitem[{Locke et~al.(2017)Locke, Zuccon, and Scells}]{locke2017automatic}
Daniel Locke, Guido Zuccon, and Harrisen Scells. 2017.
\newblock Automatic query generation from legal texts for case law retrieval.
\newblock In \emph{Information Retrieval Technology: 13th Asia Information
  Retrieval Societies Conference, AIRS 2017, Jeju Island, South Korea, November
  22-24, 2017, Proceedings 13}, pages 181--193. Springer.

\bibitem[{Ma et~al.(2022)Ma, Guo, Zhang, Fan, and Cheng}]{ma2022scattered}
Xinyu Ma, Jiafeng Guo, Ruqing Zhang, Yixing Fan, and Xueqi Cheng. 2022.
\newblock Scattered or connected? an optimized parameter-efficient tuning
  approach for information retrieval.
\newblock In \emph{Proceedings of the 31st ACM International Conference on
  Information \& Knowledge Management}, pages 1471--1480.

\bibitem[{Ma et~al.(2021)Ma, Shao, Wu, Liu, Zhang, Zhang, and
  Ma}]{ma2021lecard}
Yixiao Ma, Yunqiu Shao, Yueyue Wu, Yiqun Liu, Ruizhe Zhang, Min Zhang, and
  Shaoping Ma. 2021.
\newblock Lecard: a legal case retrieval dataset for chinese law system.
\newblock In \emph{Proceedings of the 44th international ACM SIGIR conference
  on research and development in information retrieval}, pages 2342--2348.

\bibitem[{Mahabadi et~al.(2021)Mahabadi, Ruder, Dehghani, and
  Henderson}]{mahabadi2021parameter}
Rabeeh~Karimi Mahabadi, Sebastian Ruder, Mostafa Dehghani, and James Henderson.
  2021.
\newblock Parameter-efficient multi-task fine-tuning for transformers via
  shared hypernetworks.
\newblock In \emph{Proceedings of the 59th Annual Meeting of the Association
  for Computational Linguistics and the 11th International Joint Conference on
  Natural Language Processing (Volume 1: Long Papers)}, pages 565--576.

\bibitem[{Mandal et~al.(2017)Mandal, Ghosh, Bhattacharya, Pal, and
  Ghosh}]{mandal2017overview}
Arpan Mandal, Kripabandhu Ghosh, Arnab Bhattacharya, Arindam Pal, and Saptarshi
  Ghosh. 2017.
\newblock Overview of the fire 2017 irled track: Information retrieval from
  legal documents.
\newblock In \emph{FIRE (Working Notes)}, pages 63--68.

\bibitem[{Mao et~al.(2022)Mao, Mathias, Hou, Almahairi, Ma, Han, Yih, and
  Khabsa}]{mao2022unipelt}
Yuning Mao, Lambert Mathias, Rui Hou, Amjad Almahairi, Hao Ma, Jiawei Han,
  Scott Yih, and Madian Khabsa. 2022.
\newblock Unipelt: A unified framework for parameter-efficient language model
  tuning.
\newblock In \emph{Proceedings of the 60th Annual Meeting of the Association
  for Computational Linguistics (Volume 1: Long Papers)}, pages 6253--6264.

\bibitem[{Mochales and Moens(2008)}]{mochales2008study}
Raquel Mochales and Marie-Francine Moens. 2008.
\newblock Study on the structure of argumentation in case law.
\newblock In \emph{Proceedings of the 2008 conference on legal knowledge and
  information systems}, pages 11--20.

\bibitem[{Moens(2001)}]{moens2001innovative}
Marie-Francine Moens. 2001.
\newblock Innovative techniques for legal text retrieval.
\newblock \emph{Artificial Intelligence and Law}, 9:29--57.

\bibitem[{Navas-Loro and Rodriguez-Doncel(2022)}]{navas2022whenthefact}
Mar{\i}a Navas-Loro and V{\i}ctor Rodriguez-Doncel. 2022.
\newblock Whenthefact: Extracting events from european legal decisions.
\newblock In \emph{Legal Knowledge and Information Systems}, pages 219--224.
  IOS Press.

\bibitem[{Pal et~al.(2023)Pal, Lassance, D{\'e}jean, and
  Clinchant}]{pal2023parameter}
Vaishali Pal, Carlos Lassance, Herv{\'e} D{\'e}jean, and St{\'e}phane
  Clinchant. 2023.
\newblock Parameter-efficient sparse retrievers and rerankers using adapters.
\newblock In \emph{European Conference on Information Retrieval}, pages 16--31.
  Springer.

\bibitem[{Paul et~al.(2022)Paul, Goyal, and Ghosh}]{paul2022lesicin}
Shounak Paul, Pawan Goyal, and Saptarshi Ghosh. 2022.
\newblock Lesicin: A heterogeneous graph-based approach for automatic legal
  statute identification from indian legal documents.
\newblock In \emph{Proceedings of the AAAI conference on artificial
  intelligence}, volume~36, pages 11139--11146.

\bibitem[{Pfeiffer et~al.(2021)Pfeiffer, Kamath, R{\"u}ckl{\'e}, Cho, and
  Gurevych}]{pfeiffer2021adapterfusion}
Jonas Pfeiffer, Aishwarya Kamath, Andreas R{\"u}ckl{\'e}, Kyunghyun Cho, and
  Iryna Gurevych. 2021.
\newblock Adapterfusion: Non-destructive task composition for transfer
  learning.
\newblock In \emph{Proceedings of the 16th Conference of the European Chapter
  of the Association for Computational Linguistics: Main Volume}, pages
  487--503.

\bibitem[{Piroi et~al.(2013)Piroi, Lupu, and Hanbury}]{piroi2013overview}
Florina Piroi, Mihai Lupu, and Allan Hanbury. 2013.
\newblock Overview of clef-ip 2013 lab: Information retrieval in the patent
  domain.
\newblock In \emph{Information Access Evaluation. Multilinguality,
  Multimodality, and Visualization: 4th International Conference of the CLEF
  Initiative, CLEF 2013, Valencia, Spain, September 23-26, 2013. Proceedings
  4}, pages 232--249. Springer.

\bibitem[{Poje(2014)}]{poje2014legal}
Joshua Poje. 2014.
\newblock Legal research.
\newblock \emph{American Bar Association Techreport}, 2014.

\bibitem[{Poudyal et~al.(2019)Poudyal, Gon{\c{c}}alves, and
  Quaresma}]{poudyal2019using}
Prakash Poudyal, Teresa Gon{\c{c}}alves, and Paulo Quaresma. 2019.
\newblock Using clustering techniques to identify arguments in legal documents.
\newblock \emph{ASAIL@ ICAIL}, 2385.

\bibitem[{Poudyal et~al.(2020)Poudyal, {\v{S}}avelka, Ieven, Moens, Goncalves,
  and Quaresma}]{poudyal2020echr}
Prakash Poudyal, Jarom{\'\i}r {\v{S}}avelka, Aagje Ieven, Marie~Francine Moens,
  Teresa Goncalves, and Paulo Quaresma. 2020.
\newblock Echr: Legal corpus for argument mining.
\newblock In \emph{Proceedings of the 7th Workshop on Argument Mining}, pages
  67--75.

\bibitem[{Rabelo et~al.(2022)Rabelo, Goebel, Kim, Kano, Yoshioka, and
  Satoh}]{rabelo2022overview}
Juliano Rabelo, Randy Goebel, Mi-Young Kim, Yoshinobu Kano, Masaharu Yoshioka,
  and Ken Satoh. 2022.
\newblock Overview and discussion of the competition on legal information
  extraction/entailment (coliee) 2021.
\newblock \emph{The Review of Socionetwork Strategies}, 16(1):111--133.

\bibitem[{Robertson et~al.(1995)Robertson, Walker, Jones, Hancock-Beaulieu,
  Gatford et~al.}]{robertson1995okapi}
Stephen~E Robertson, Steve Walker, Susan Jones, Micheline~M Hancock-Beaulieu,
  Mike Gatford, et~al. 1995.
\newblock Okapi at trec-3.
\newblock \emph{Nist Special Publication Sp}, 109:109.

\bibitem[{Ruder et~al.(2022)Ruder, Pfeiffer, and Vuli{\'c}}]{ruder2022modular}
Sebastian Ruder, Jonas Pfeiffer, and Ivan Vuli{\'c}. 2022.
\newblock Modular and parameter-efficient fine-tuning for nlp models.
\newblock In \emph{Proceedings of the 2022 Conference on Empirical Methods in
  Natural Language Processing: Tutorial Abstracts}, pages 23--29.

\bibitem[{Sansone and Sperl{\'\i}(2022)}]{sansone2022legal}
Carlo Sansone and Giancarlo Sperl{\'\i}. 2022.
\newblock Legal information retrieval systems: State-of-the-art and open
  issues.
\newblock \emph{Information Systems}, 106:101967.

\bibitem[{Santosh et~al.(2023)Santosh, Ichim, and Grabmair}]{tyss2023zero}
Tyss Santosh, Oana Ichim, and Matthias Grabmair. 2023.
\newblock Zero-shot transfer of article-aware legal outcome classification for
  european court of human rights cases.
\newblock In \emph{Findings of the Association for Computational Linguistics:
  EACL 2023}, pages 593--605.

\bibitem[{Santosh et~al.(2022)Santosh, Xu, Ichim, and
  Grabmair}]{santosh2022deconfounding}
Tyss Santosh, Shanshan Xu, Oana Ichim, and Matthias Grabmair. 2022.
\newblock Deconfounding legal judgment prediction for european court of human
  rights cases towards better alignment with experts.
\newblock In \emph{Proceedings of the 2022 Conference on Empirical Methods in
  Natural Language Processing}, pages 1120--1138.

\bibitem[{Shukla et~al.(2022)Shukla, Bhattacharya, Poddar, Mukherjee, Ghosh,
  Goyal, and Ghosh}]{shukla2022legal}
Abhay Shukla, Paheli Bhattacharya, Soham Poddar, Rajdeep Mukherjee, Kripabandhu
  Ghosh, Pawan Goyal, and Saptarshi Ghosh. 2022.
\newblock Legal case document summarization: Extractive and abstractive methods
  and their evaluation.
\newblock In \emph{Proceedings of the 2nd Conference of the Asia-Pacific
  Chapter of the Association for Computational Linguistics and the 12th
  International Joint Conference on Natural Language Processing}, pages
  1048--1064.

\bibitem[{Singh et~al.(2021)Singh, Reddy, Hamilton, Dyer, and
  Yogatama}]{singh2021end}
Devendra Singh, Siva Reddy, Will Hamilton, Chris Dyer, and Dani Yogatama. 2021.
\newblock End-to-end training of multi-document reader and retriever for
  open-domain question answering.
\newblock \emph{Advances in Neural Information Processing Systems},
  34:25968--25981.

\bibitem[{Tam et~al.(2022)Tam, Liu, Ji, Xue, Zhang, Dong, Liu, Hu, and
  Tang}]{tam2022parameter}
Weng~Lam Tam, Xiao Liu, Kaixuan Ji, Lilong Xue, Xingjian Zhang, Yuxiao Dong,
  Jiahua Liu, Maodi Hu, and Jie Tang. 2022.
\newblock Parameter-efficient prompt tuning makes generalized and calibrated
  neural text retrievers.
\newblock \emph{arXiv preprint arXiv:2207.07087}.

\bibitem[{Thakur et~al.(2021)Thakur, Reimers, R{\"u}ckl{\'e}, Srivastava, and
  Gurevych}]{thakur2021beir}
Nandan Thakur, Nils Reimers, Andreas R{\"u}ckl{\'e}, Abhishek Srivastava, and
  Iryna Gurevych. 2021.
\newblock Beir: A heterogeneous benchmark for zero-shot evaluation of
  information retrieval models.
\newblock In \emph{Thirty-fifth Conference on Neural Information Processing
  Systems Datasets and Benchmarks Track (Round 2)}.

\bibitem[{Tyss et~al.(2023)Tyss, San~Blas, Kemper, and
  Grabmair}]{tyss2023leveraging}
Santosh Tyss, Marcel~Perez San~Blas, Phillip Kemper, and Matthias Grabmair.
  2023.
\newblock Leveraging task dependency and contrastive learning for case outcome
  classification on european court of human rights cases.
\newblock In \emph{Proceedings of the 17th Conference of the European Chapter
  of the Association for Computational Linguistics}, pages 1103--1103.

\bibitem[{Verma et~al.(2020)Verma, Morato, Jain, and Arora}]{verma2020relevant}
Aayushi Verma, Jorge Morato, Arti Jain, and Anuja Arora. 2020.
\newblock Relevant subsection retrieval for law domain question answer system.
\newblock \emph{Data Visualization and Knowledge Engineering: Spotting Data
  Points with Artificial Intelligence}, pages 299--319.

\bibitem[{Wang et~al.(2018)Wang, Yang, Niu, Zhang, Zhang, and
  Niu}]{wang2018modeling}
Pengfei Wang, Ze~Yang, Shuzi Niu, Yongfeng Zhang, Lei Zhang, and ShaoZhang Niu.
  2018.
\newblock Modeling dynamic pairwise attention for crime classification over
  legal articles.
\newblock In \emph{the 41st international ACM SIGIR conference on research \&
  development in information retrieval}, pages 485--494.

\bibitem[{Xiong et~al.(2020)Xiong, Xiong, Li, Tang, Liu, Bennett, Ahmed, and
  Overwijk}]{xiong2020approximate}
Lee Xiong, Chenyan Xiong, Ye~Li, Kwok-Fung Tang, Jialin Liu, Paul~N Bennett,
  Junaid Ahmed, and Arnold Overwijk. 2020.
\newblock Approximate nearest neighbor negative contrastive learning for dense
  text retrieval.
\newblock In \emph{International Conference on Learning Representations}.

\bibitem[{Xu et~al.(2023{\natexlab{a}})Xu, Staufer, T.y.s.s, Ichim, Heri, and
  Grabmair}]{xu2023vechr}
Shanshan Xu, Leon Staufer, Santosh T.y.s.s, Oana Ichim, Corina Heri, and
  Matthias Grabmair. 2023{\natexlab{a}}.
\newblock {VECHR}: A dataset for explainable and robust classification of
  vulnerability type in the {E}uropean court of human rights.
\newblock In \emph{Proceedings of the 2023 Conference on Empirical Methods in
  Natural Language Processing}, pages 11738--11752, Singapore. Association for
  Computational Linguistics.

\bibitem[{Xu et~al.(2023{\natexlab{b}})Xu, T.y.s.s, Ichim, Risini, Plank, and
  Grabmair}]{xu2023dissonance}
Shanshan Xu, Santosh T.y.s.s, Oana Ichim, Isabella Risini, Barbara Plank, and
  Matthias Grabmair. 2023{\natexlab{b}}.
\newblock From dissonance to insights: Dissecting disagreements in rationale
  construction for case outcome classification.
\newblock In \emph{Proceedings of the 2023 Conference on Empirical Methods in
  Natural Language Processing}, pages 9558--9576, Singapore. Association for
  Computational Linguistics.

\bibitem[{Yates et~al.(2021)Yates, Nogueira, and Lin}]{yates2021pretrained}
Andrew Yates, Rodrigo Nogueira, and Jimmy Lin. 2021.
\newblock Pretrained transformers for text ranking: Bert and beyond.
\newblock In \emph{Proceedings of the 14th ACM International Conference on web
  search and data mining}, pages 1154--1156.

\bibitem[{Zhang et~al.(2020)Zhang, Sax, Zamir, Guibas, and
  Malik}]{zhang2020side}
Jeffrey~O Zhang, Alexander Sax, Amir Zamir, Leonidas Guibas, and Jitendra
  Malik. 2020.
\newblock Side-tuning: a baseline for network adaptation via additive side
  networks.
\newblock In \emph{Computer Vision--ECCV 2020: 16th European Conference,
  Glasgow, UK, August 23--28, 2020, Proceedings, Part III 16}, pages 698--714.
  Springer.

\end{thebibliography}
